\documentclass[twoside,11pt]{article}

\usepackage{blindtext}
\pdfoutput=1

\usepackage{jmlr2e}
\usepackage{slashbox}
\usepackage[utf8]{inputenc} 
\usepackage[T1]{fontenc}    
\usepackage{hyperref}       
\usepackage{url}            
\usepackage{booktabs}       
\usepackage{amsfonts}       
\usepackage{nicefrac}       
\usepackage{microtype}      
\usepackage{xcolor}         
\usepackage{amsmath}
\usepackage{comment}
\usepackage{algorithm}
\usepackage{algpseudocode}
\usepackage{graphicx}
\PassOptionsToPackage{square}{natbib} 
\usepackage{jmlr2e}
\usepackage{subfigure}

\usepackage{lastpage}
\jmlrheading{23}{2022}{1-\pageref{LastPage}}{1/21; Revised 5/22}{9/22}{21-0000}{Hugo Laurencon and Yesoda Bhargava}

\ShortHeadings{Continuous Time Continuous Space Homeostatic Reinforcement Learning (CTCS-HRRL)}{}
\firstpageno{1}

\begin{document}

\title{Continuous Time Continuous Space Homeostatic Reinforcement Learning (CTCS-HRRL) : Towards Biological Self-Autonomous Agent}

\author{\name Hugo Laurencon$^\star$ \email hugo.laurencon@gmail.com \\
       \addr Group for Neural Theory, Ecole Normale Superieure, \\
       Paris, France\\
       \AND
       \name Yesoda Bhargava$^\star$ \email yesodabhargava@gmail.com \\
       \addr Cognitive Neuroscience Lab\\
      BITS Pilani K. K. Birla Goa Campus\\
       Goa, India \\
       \AND
       \name Riddhi Zantye \email riddhi.zantye@gmail.com \\
       \addr Cognitive Neuroscience Lab\\
      BITS Pilani K. K. Birla Goa Campus\\
       Goa, India \\
       \AND
       \name Charbel-Raphaël Ségerie \email crsegerie@gmail.com \\
       \addr Ecole Normale Supérieure Paris-Saclay,\\
	Université Paris-Saclay,\\
       Paris, France\\
       \AND
       \name Johann Lussange \email  johann.lussange@ens.fr \\
       \addr Group for Neural Theory, Ecole Normale Superieure, \\
       Paris, France\\
       \AND 
       \name Veeky Baths \email veeky@goa.bits-pilani.ac.in \\
       \addr Cognitive Neuroscience Lab\\
      BITS Pilani K. K. Birla Goa Campus\\
       Goa, India \\
       \AND
       \name Boris Gutkin*\email boris.gutkin@ens.fr \\
       \addr Group for Neural Theory, Ecole Normale Superieure, \\
       Paris, France}

\editor{}

\maketitle
\def\thefootnote{$^\star$}\footnotetext{Equal Contribution.}\def\thefootnote{\arabic{footnote}}
\def\thefootnote{*}\footnotetext{Corresponding author}\def\thefootnote{\arabic{footnote}}
\newpage
\begin{abstract}
Homeostasis is a biological process by which living beings maintain their internal balance. Previous research suggests that homeostasis is a learned behaviour. Recently introduced Homeostatic Regulated Reinforcement Learning (HRRL) framework attempts to explain this learned homeostatic behavior by linking Drive Reduction Theory and Reinforcement Learning. This linkage has been proven in the discrete time-space, but not in the continuous time-space. In this work, we advance the HRRL framework to a continuous time-space environment and validate the CTCS-HRRL (Continuous Time Continuous Space HRRL) framework. We achieve this by designing a model that mimics the homeostatic mechanisms in a real-world biological agent. This model uses the Hamilton-Jacobian Bellman Equation, and function approximation based on neural networks and Reinforcement Learning. Through a simulation-based experiment we demonstrate the efficacy of this model and uncover the evidence linked to the agent's ability to dynamically choose policies that favor homeostasis in a continuously changing internal-state milieu. Results of our experiments demonstrate that agent learns homeostatic behaviour in a CTCS environment, making CTCS-HRRL a promising framework for modellng animal dynamics and decision-making. 
\end{abstract}


\section{Introduction}
Reinforcement learning (RL) has been of particular interest in recent years in the area of Machine Learning (ML) and Artificial Intelligence (AI). Dramatic advances have been made \citep{Mnih, Silver, Silver2018} particularly due to the progress in Deep Learning (DL) \citep{Krizhevsky}. These advances are also due to the easy applicability of the general RL framework to many fields, such as Economics \citep{Lussange}, Psychology \citep{Shteingart}, Control Theory \citep{Kretchmar} and Neuroscience \citep{Niv}. Additionally, the intermingling of Neuroscience and AI has further advanced the applicability of RL to real-world problems \citep{Kriegeskorte, Richards}. 

A natural next step to these advancements is designing self-autonomous agents that may mimic behaviour of the real-world biological and psychological agents (e.g. rodents, primates, humans). More specifically, the overarching goal is to develop agents that can express both physiological and psychological needs akin to biological organisms and are capable of learning, acting and adapting in a given environment depending upon their internal states and the external environment. In this context, homeostatic and allostatic regulation principles are relevant. Following these principles, \citeauthor{Man} defined a class of robots capable of exhibiting emotions, and equipped with the ability to learn and adapt in unfamiliar environments while simultaneously observing their internal states \citep{Man}. 

Computational integration of these homeostatic and allostatic principles in the agent can be achieved using RL methodologies. However, RL alone proves insufficient as it seeks to maximise the rewards based on action. Whereas, in the case of bio-mimetic agents, the requirement is to learn both reward maximization and homeostatic deviation minimization \citep{Staddon, Toates}. In this context, the Homeostatic Regulation Theory (HRT) is particularly relevant \citep{Keramati}. While RL and HRT may seem divergent, \citeauthor{Keramati} linked the two theories by demonstrating that the goal of reward maximization (RL) and homeostatic deviation minimization (HRT) is equivalent when the reward function is based on the internal state of the agent. This compound theory proposed by \citeauthor{Keramati} is now known as the Homeostatically Regulated Reinforcement Learning (HRRL) and lends feasibility to the development of bio-mimetic agents. In fact, HRRL has been found to effectively model primitive behaviours such as resource consumption, and evolved behaviours such as risk aversion, alcohol tolerance, cocaine addiction or anticipatory control \citep{Keramati2014, KeramatiThesis}. Thus, RL and Homeostatic Regulation Theory together provide a more robust and practical mechanism to develop self-autonomous bio-mimetic agents. 

Despite the feasibility that HRRL lends in self-autonomous agent development, it has some limitations. According to the current HRRL framework, the internal state of the agent is fixed when it is in an inactive state. Moreover, in the current HRRL framework, homeostasis is considered an episodic event rather than a continuous goal. Thus, the traditional HRRL theory does not consider the possibility of homeostatic deviation even when the agent is in an inactive state. Naturally, such a stance is not compatible with the actual behaviour of the real-world biological agents. In fact, biological agents continuously monitor their internal state and are aware of homeostatic deviation that may result due to the internal processes required for survival and sustenance. Thus, a threat to homeostatic balance is actively present and biological agents continuously pursue homeostatic regulation which involves physiological or behavioural change \citep{Ramsay}. 

A second limitation of the current HRRL framework is  the discrete mapping of action and time \citep{Keramati}, i.e.  the actions taken by the agent are assumed to be carried out at discrete and regular time steps. Whereas in the real-world actions are generally carried out in a continuous and smooth manner. Moreover, the current HRRL framework is based on a discount factor which does not model the notion of temporality between actions. 

In our present work, we aim to address the above explained shortcomings of the current HRRL framework. To this end, we advance the HRRL framework to the continuous-time and continuous-space (CTCS) paradigm, thereby formulating the CTCS-HRRL model. The main contributions of our work are: 

\begin{itemize}
	\item \textbf{Dynamic Self-Regulating Agents:} Homeostatic behaviour is embodied within the agent irrespective of its state (inactive or active). This embodiment is based on real-world observations. Psychological and behavioural attributes are also embodied to emulate the real-world biological agents. These behaviours are sleeping, resting, walking instead of running. Thus, agent's self-regulation is guided by an active knowledge and awareness of its internal states. 
	\item \textbf{Continuous time implementation:} Unlike the previous learning models that focused on discrete-time learning, we extend these models by introducing the continuous-time learning framework in the HRRL. We also demonstrate the transferability of the theoretical results in discrete model to the continuous model. 	
	\item \textbf{Agent-Environment Interaction and Self-Learning Agent:} Limited research currently exists on the role of agent-environment interaction in agent's decision-making. We embed the agent-environment interaction in our simulation experiments to mimic the decision making of the real-world biological agents. Due to this, the agent learns policies which are more realistic and ecologically valid. 
	
\end{itemize}

\section{Background and Related Work}
In this section, we critically discuss the scientific works which have attempted to incorporate agent's internal state dynamic into learning and motivation. We subsequently place our work in the context of the discussion. 

\textbf{Negative Feedback Models}: These models relate to the control theory and formulate homeostatic deviation as a negative feedback state. This homeostatic deviation indicates the drive of the organism towards a particular resource, and is assumed to guide the behaviour of the organism. Thus, in these models, the behavior of the organism is dependent on its internal state only. Greater the homeostatic deviation, greater the motivational drive to fulfil the need. Heuristically, the negative feedback models only measure the organism's discomfort or the negative effect. However, behaviour selection or prioritization is not discussed. 

\textbf{Drive Reduction Theory}:  A natural extension of the negative feedback models is the Drive Reduction Theory proposed by Hull \citep{Hull}. According to the theory, the organism selects actions (behaviours) to reduce its drive, or the homeostatic deviation. Although DRT has explained the adaptive behaviour, learning \citep{Staddon} and motivational systems \citep{Toates}, it fails to explain the conceptual and mathematical basis of action selection. Furthermore, it does not explain the peculiar behaviour of resource consumption in the absence of homeostatic deviation \citep{Wingfield}. Specifically, the behaviour taken in the anticipation of a future perceived need or perceived homeostatic deviation, such as anticipatory consumption, is not explained by DRT. This gap is crucial to address because in the real world, such behaviour (anticipatory consumption or response) are common. E.g. Overconsumption of food, addiction etc. This drawback motivate the formulation of a theory that can provide a more ecologically valid explanation for anticipatory or compulsive behaviours, while rooted in homeostatic regulation. Hullian drives address this drawback. 

\textbf{Hullian Drives}: A Hullian drive is a drive that varies between 0 and 1. The 0 denotes total dissatisfaction and 1 denotes total satisfaction. Hullian drive has been used to explain the agent's behaviour and motivation in reinforcement learning based settings. For example, \citeauthor{Konidaris} used Hullian drive-based reward model weighted by the time-dependent coefficients to indicate the drive priority. However, external information on drive priority is counter-intuitive and incompatible with the intelligence of real-world biological agents, as they are able to discern these priorities automatically through an internal mechanism.  Thus, in our work we do not externally provide the information of drive priority to the agent, instead let the agent learn that on its own and accordingly modify its policies. Secondly, in the work by \citeauthor{Konidaris}, the agent is penalized if its actions do not follow the drive priority. We instead achieve this regulatory effect using a function of time and control that accounts for correlations between different drives, and assist the agent to take decisions accordingly. A third drawback of the work using Hullian drive \citep{Konidaris} is the use of SARSA algorithm which is not always robust for small time-steps, a necessity for agent learning in an unknown environment. This limitation is addressed in our study.

\textbf{Homeostatically Regulated RL (HRRL or HRL):} In addition to the Negative Feedback Models, Drive Reduction Theory, Hullian Drive, we also discuss the HRRL. According to HRRL, agent's drive trigger homeostasis-ensuring actions. The selection of actions is guided by reinforcement learning framework. Thus, in HRRL the rewards and punishments are derived directly from the internal state deviations and the Hullian drive function \citep{Keramati}. The similarity between Homeostasis achievement and Reinforcement Learning was proved when \citeauthor{Keramati} showed that maximization of the sum of discounted rewards (RL) is equivalent to the minimization of the sum of discounted drives (homeostasis). Thus, a conceptually robust reinforcement learning framework for drive reduction theory was produced which addressed the gaps in the earlier models. Although theoretical and mathematical results for HRRL have been achieved, the numerical or computer-based simulations that support these theoretical results are lacking. In this work, we advance the HRRL theory by performing numerical simulations using an artificial agent in an unknown environment. 

In the next sections we discuss the methods, experimental set up, theoretical and simulation based results. 

\section{Methods}
According to the general reinforcement learning framework \citep{Sutton}, an agent, in a certain state, selects actions from a set of available actions in that state. The choice of the action changes the agent's current state and confers it a reward (either positive or negative). A series of such actions at each state-time $t$ that maximise the discounted sum of future rewards constitute a \textit{policy}. Ultimately, the task of the agent is to discover this policy for a given task in a particular environment. In our work, we have an agent in a square environment in which it has to consume the resources as per its need (internal state) and maintain homeostasis. For this task, we use certain notations that we discuss next. 

\begin{figure}[H]
	\centering
	\includegraphics[width=6cm]{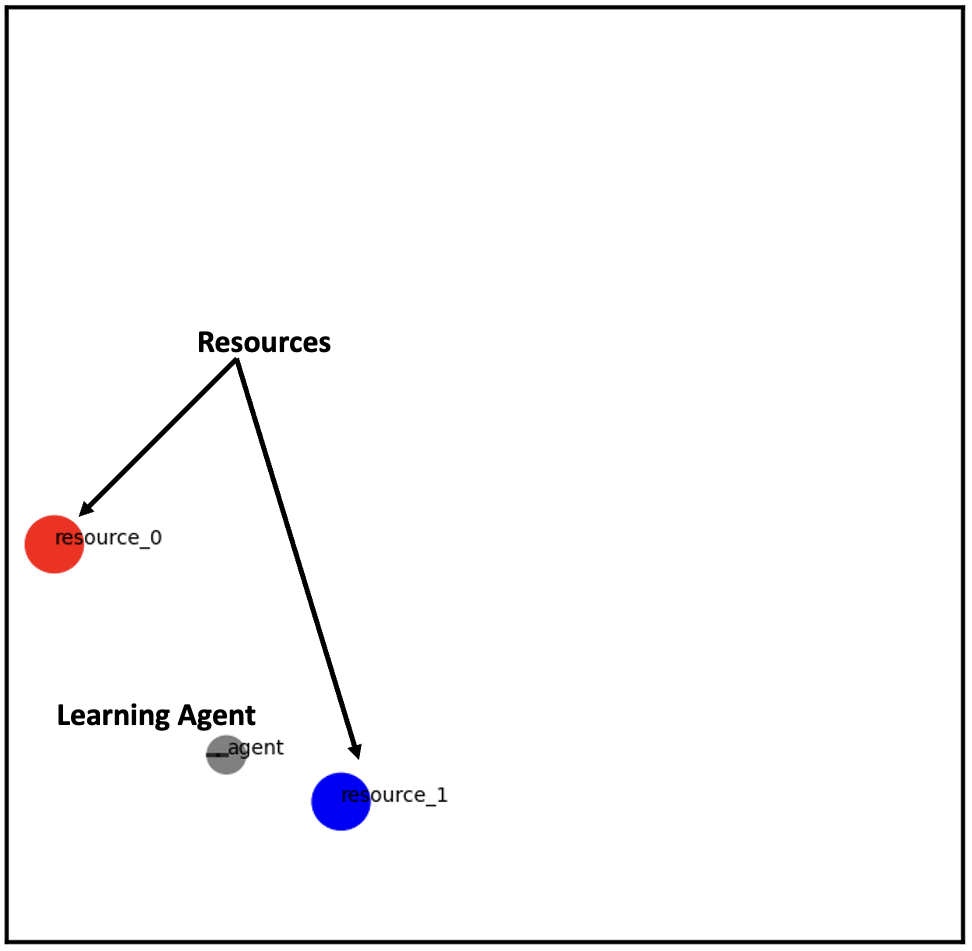}
	\caption{The environment of the simulation experiment. The agent is represented by a gray point and is located by its coordinates in the plane. The colored circles indicate the two resources present in the environment that the agent has to consume. These colored circles delimit the space in which it is possible to consume a resource. }
	\label{fig:environment}
\end{figure}

Let $n$ be an integer, $\zeta_t \in \mathrm{R}^n$ the state (internal and external) of the agent at time $t$ and $\zeta: t \mapsto \zeta_t$ the trajectory function of the agent's state. We denote the space of possible actions at time $t$ when the agent is in $\zeta_t$ by $\mathcal{A}_{\zeta_t, t}$ and the space of all actions by $\mathcal{A}$. The policy function determining the agent's choices is denoted by $\pi: \mathrm{R}^{n} \mathrm{R} \to \mathcal{A}_{\zeta_t, t}, (\zeta_t, t) \mapsto a$, and the reward received by following this policy at time $t$ by $r(t) = r_{\pi(\zeta_t, t), \zeta_t, t}$. We denote by $\Pi$ the set of all admissible policies. Note that in the deterministic case and for a fixed policy, the initial state $(\zeta_0, t_0)$ and the policy function $\pi$ completely determine the path function $\zeta$ and the reward function $r$. The value function, for an agent in $\zeta_t$ at time $t$ following a policy $\pi$, is defined as follows
\begin{equation}
	V^\pi(\zeta_t, t) = \int_{t}^{\infty} \gamma^{s-t} r(s) \, \mathrm{d}s \label{eq:V}
\end{equation}
where $\gamma \in ]0,1[$ is the discount factor that accounts for time preference. The optimization problem for the agent is
\begin{equation}
		\operatorname*{arg max}_{\pi \in \Pi} V^\pi, \label{eq:problem}
\end{equation}
where it is trying to find the policy function that maximizes the value function for each state at each time.

\paragraph{Internal Environment}
Let $x \in \mathrm{R}^{n_{int}}$ be a vector describing agent's internal state where each feature needs to be regulated, and let $x^*$ denote its homeostatic set point. Assume that each feature of $x$ is bounded. This is a constraint from the agent's embodiment, which will not be valid if a feature is too small or too large, and will either be regulated automatically by the organism or will cause the agent's death. We assume that the agent knows the differences between the individual internal variables and their respective set points: $\delta := x - x^*$, where the order of the difference is arbitrarily chosen and can be evaluated as RL comparison rewards \citep{Matignon}. 

\paragraph{External Environment}We also define the external environment of the agent within its view-field $e \in \mathrm{R}^{n_{ext}}$, and the entire world $\zeta = [\delta^{T}, e^{T}]^{T}$. At any time $t$, the agent in $\zeta_t$ can perform an action $a \in \mathcal{A}_{\zeta_t, t}$. The agent is limited in its choice by its environment (for example by the place in which it is) and by its internal state and time (because some actions depend on energy). The action taken, in turn, will have a consequence (a control $u \in \mathrm{R}^{n_{int} + n_{ext}}$) for its internal state and its environment. We assume that from the agent's point of view, the dynamics of $\zeta$ is described by an equation of the form
\begin{equation}
	d \zeta = f(\zeta, u, t)dt + g(\zeta, t)dS \label{eq:main}
\end{equation}
where $f, g$ are functions and $S$ is a stochastic process. $f$, $g$ and $S$ are unknown to the agent at the beginning of its task: it does not have the information of how its body and the external world react and has no estimate of the behavior of the stochastic process $S$.

We can distinguish several potentially overlapping causes of a change in the agent's internal state and environment:
\newline
(a) an unconscious automatic autoregulation of the organism, modeled by the function $f$ and its variable $\zeta$, which can account for internal processes of the agent's body (e.g., animal physiology, robot mechanics, and multi-component interactions). We note that from the perspective of biology, self-regulation is a common physiological process \citep{Polynikis,Pattaranit}. For e.g. the human kidney uses a mechanism called tubuloglomerular feedback to regulate the glomerular filtration rate in response to changes in sodium concentration \citep{Versypt,Thomson}. Autoregulation of cerebral blood flow has been demonstrated in the presence of Carbon Dioxide \citep{Panerai}. This change in $\zeta$ takes place without the agent taking any action of its own. If the agent does not perform any action, its internal state will naturally deviate from its current state;
\newline
(b) a control that the agent exercises and that has an impact on its environment and its internal state, modeled by the function $f$ and its variable $u$. For example, the action of moving, which modifies the environment while requiring an effort, and which thus has an impact on the internal state;
\newline
(c) the time that changes and modifies the external environment and the internal state, modeled by the function $f$ and its variable $t$. For example, the time of day and the current season will drive the temperature and brightness of the environment in a certain direction. Time can also model old age, by progressively modifying the function $f$, and thus the way the body reacts over time;
\newline
(d) a stochastic control that the agent undergoes, leading to an unexpected change mostly in the environment, modeled by the $g$ function and the stochastic process $S$. Stochasticity intervenes in everything that the agent cannot control, in particular the behavior of other agents around it or the weather.

The control, by changing the environment, has a direct impact on the actions that can be taken in the future, since these will depend on the new environment. By changing the internal state, it also has a direct impact on the agent's drive $d$. Control brings the agent to a more or less comfortable state, depending on whether it is moving towards or away from its homeostatic set point $x^*$, or equivalently whether the drive is decreasing or increasing.

\paragraph{Agent's Goal}The agent's goal is to minimize its drive for the two resources by finding the optimal policy that allows it to take actions aligned with the achievement of homeostasis. The deviation function $J^\pi : \mathrm{R}^{n_{int} + n_{ext}} \times \mathrm{R} \to \mathrm{R}$ for an admissible policy $\pi$, which represents the integral of the agent's discounted drive over its remaining lifetime, is given by
\begin{equation}
	J^\pi(\zeta_t, t) = \mathbb{E} \left ( \int_{t}^{\infty} \gamma^{s-t} d(\delta(s)) \mathrm{d}s \right ) \ \label{eq:J}
\end{equation}
where $\zeta$ follows equation \eqref{eq:main} (and thus $\delta$ depends on $\pi$) on $[t,+\infty[$ with the initial condition $\zeta(t) = \zeta_t$, the control function $u$ satisfies $\forall s \in [t,+\infty[, u(s) = u_{\pi(\zeta(s),s)}$ (we will say that the control function is associated with the policy if it meets this last condition) and the expected value is here because of the stochasticity in \eqref{eq:main}. Note that the integral is well-defined thanks to the discount factor and the fact that $x$ and $x^*$ are bounded. Concretely, the value of the deviation function $J^\pi(\zeta_t, t)$ indicates how bad it is for the agent to follow the policy $\pi$, starting from the state $\zeta_t$ at time $t$. The problem of the agent is thus : 
\begin{equation}
	\operatorname*{arg min}_{\pi \in \Pi} J^\pi \label{eq:J_problem}
\end{equation}
with the same conditions as before. At a given time $t$, $\zeta_t$ values are continuous, action $a_t$ values are discrete and the associated control $u_t$ is generally small. 

\paragraph{Hamilton-Jacobi Bellman Equation}
\begin{equation}
	-\log(\gamma)J^{*}(\zeta_t) = \min_{a \in \mathcal{A}_{\zeta_t}} d(\zeta_t, u_a) + \frac{\partial J^{*}}{\partial \zeta}(\zeta_t) \cdot f(\zeta_t, u_a)
	\label{eq: hjb}
\end{equation}
where $J^{*}$ is the optimal deviation function, $u_a$ is the deterministic control resulting from the action $a$ and $d$ is the drive function, with the conventions that $d(\zeta_t, u_a)$ is the drive of the new state of the agent after performing the action $a$ in $\zeta_t$ (entire world = internal state + external environment), and that symbolically $d(\zeta_t) = d(\delta_t)$. The intuition behind this equation is obtained with the optimality principle and by making the analogy with the known discrete Bellman equation \citep{Sutton}. Because $Q$-learning is not robust in the presence of small-time steps\citep{Tallec}, we rely on this equation to propose our algorithm.

\subsection{Experiment}

\subsubsection{Description of the experiment}
We consider a closed 2D environment (Figure~\ref{fig:environment}). The agent is identified by its coordinates in the plane (grey patch in  Figure~\ref{fig:environment}). The environment contains hidden resources necessary for the agent's survival. For the sake of this experiment, we have two stationary resources, blue and green as shown in Figure~\ref{fig:environment}.  For a biological agent, these resources could act as a reservoir of proteins, a source of carbohydrates, water, or any other element crucial for their survival. The agent's internal state is determined by these resources in its system. 

For the simulation purposes, we used a square environment with two resources ($R_1 \text{ and } R_2$) and one single agent. The agent's starting internal state for each resource is very minimal, but not so low as to cause muscular fatigue and prevent movement. The homeostatic set points are $R_1=1$ and $R_2=2$. Thus, the task of the agent is to maintain homeostasis (minimise the deviation function $J$) over changing internal states. The possible actions for the agent are: walk, run, go to the resource, consume resource, and rest in case of excess fatigue (muscular or sleep-related). At each instant, the agent can move forward by an elementary distance for the action of walking (up, down, right, or left) or by a greater elementary distance for the action of running (only when the agent is near the resource). 

In addition to certain actions that an agent can take, we endow the agent with physiological properties (internal state or body dynamics including fatigue and the state of immobilized sleep). The model of the internal state (the body) of the agent includes two types of fatigue, "muscle" fatigue, which depends on how far the agent has moved without resting (e.g., continuous movement), and "sleep" fatigue, if it has not recovered for too long. Splitting fatigue into such two separate terms, allowed us to reflect multiple behavioral and physiological aspects that cause natural agents (animals) to rest. However, homeostatic state is dependent only on the concentration of resources, and not muscle or sleep fatigue. At any time and in any place, the agent can choose the action of sleeping for a minimal renewable duration. This action will immobilize it for a certain duration. 

When the agent has reached a certain threshold of muscular fatigue, it cannot take action of running, and at other threshold, it cannot walk. These thresholds are pre-decided and incorporated in the code. Such threshold based conditions ensure that the agent is immobile to recover from the muscular fatigue. Similarly beyond a certain threshold of sleep-related fatigue, the only action that becomes possible is sleeping. Thus, our agent mimics the natural biological agent.  In this environment, the agent begin with zero knowledge and its goal is to minimise the deviation function ($J$). The agent explores the environment and eventually learns to base its action exploiting previous actions. Agent has access to $\zeta$ only and based on its actions accrues rewards, and updates the deviation function. 

We run the simulation for 6000, 8000, 10000, and 14000 iterations to study the agent's learning behaviour. Since the biological process of homeostasis is continuous and never ending, the program/simulation never really ends. But a saturation stage can be noticed which establishes that the agent has thoroughly learned about resource positions and directly reaches to those reservoirs in times of internal drive or homeostatic deviation. Next  we present the learning algorithm for the agent. 

\subsection{Learning algorithm for the agent}

Here we present the Algorithm~\ref{learning_algo} that allows the agent to learn by interacting with its environment. The algorithm is based on the principle of policy improvement, wherein at each step a value function is evaluated, and the policy is updated directly using this value function. The classical reinforcement learning heuristics to improve the quality of learning are deliberately not implemented here, as the goal is not to propose the most efficient algorithm possible. In contrast, our goal is to present a proof of concept that sufficiently demonstrates the possibility for an agent to learn from zero knowledge by following a natural and plausible approach to action selection and gradually learning from the accumulated experience.

\begin{algorithm}[H]
	\footnotesize
	\caption{Learning algorithm for the agent \label{algo:algo_agent}}
	\label{learning_algo}
	\begin{algorithmic}
		\State Randomly initialize the transition function $f(\zeta, u | \theta^{f})$ and the deviation function $J(\zeta | \theta^J)$ with weights $\theta^{f}$ and $\theta^{J}$
		\State Receive initial observation state $\zeta_1$
		\For{$k = 1,\dots, K$}
		\State With probability $\epsilon$ select a random action $a_k \in \mathcal{A}_{\zeta_k}$, otherwise select
		\begin{equation*}
			a_k = 	\operatorname*{arg min}_{a \in \mathcal{A}_{\zeta_k}} d(\zeta_k + f(\zeta_k, u_{a} | \theta^{f})\Delta_t) + \frac{\partial }{\partial \zeta}J(\zeta_k | \theta^{J})\cdot f(\zeta_k, u_a | \theta^{f})
		\end{equation*}
		\State Execute action $a_k$ and observe new state $\zeta_{k+1}$
		\State Update the transition function and the deviation function by performing a gradient descent step on
		\begin{align*}
			& L_f = (\zeta_{k+1} - \zeta_k - f(\zeta_k, u_{a_k} | \theta^{f})\Delta_t)^T (\zeta_{k+1} - \zeta_k - f(\zeta_k, u_{a_k} | \theta^{f})\Delta_t) \text{ with respect to } \theta^{f} \\
			& L_J = (d(\zeta_{k+1}) + \frac{\partial }{\partial \zeta}J(\zeta_k | \theta^{J}) \cdot f(\zeta_k, u_{a_k} | \theta^{f}) + \log(\gamma) J(\zeta_k | \theta^J))^2 \text{ with respect to } \theta^{J}
		\end{align*}
		\EndFor
	\end{algorithmic}
\end{algorithm}

We discretize time by $\Delta_t$ time steps. The discretization in time is necessary to build the algorithm, but the proposal of a continuous theoretical framework is justified by a better modeling, an economy of notations, or the possibility to make adaptive time steps. In the initial state, the agent does not have access to the functioning of its internal state (the body), represented by the function $f$. Over time, the agent learns to approximate this function through its experiences. We thus have a model-based algorithm, since the transitions between internal states are modeled. On the other hand, the drive function $d$ is known initially, modeling the agent's interoception. 

The agent's action is either taken randomly with probability $\epsilon$ to facilitate exploration, or based on the HJB equation and estimates of $J$ and $f$. Note that, for a certain policy $\pi$, the deviation function $J$ is defined by equation \eqref{eq:J}. However, this equation requires the calculation of and integral over the lifetime of the agent following this policy, which is impossible for the agent since it does not have access to the information of its future. Therefore, the agent maintains an estimate of $J$ instead, and updates it according to Algorithm \ref{algo:algo_agent}.

The estimated transition and deviation functions are neural networks. The estimated deviation function $J$ is updated at each step by minimizing an associated error \citep{Doya}. The gradient of the neural networks with respect to the inputs is also computed by backpropagation.

\section{Theoretical results}

\subsubsection{An equivalent formulation of the optimization problem}

We define the reward at time $t$ for an agent whose internal state follows the trajectory function $\zeta = [\delta^{T}, e^{T}]^{T}$ as follows 
\begin{equation}
	r(t) = -(d(\delta))'(t).
	\label{eq:r}
\end{equation}
Intuitively, the reward, which can be positive or negative, is proportional to the variation of the drive of the agent, and thus to what the agent has gained (or lost) in comfort with respect to the stasis point between time $t$ and $t+dt$. This variation of the drive is implied by the control and thus by the action that the agent has taken at time $t$.

\begin{lemma}
	The pursuit of homeostatic stability is equivalent to the maximization of the reward. Formally, we have
	\begin{equation}
	\operatorname*{arg max}_{\pi \in \Pi} V^\pi = \operatorname*{arg min}_{\pi \in \Pi} J^\pi\label{eq:equivalence_second_problem}
	\end{equation}
\end{lemma}

\begin{proof}
	On doing an integration by parts (valid even in the case where the function $\zeta$ is continuous everywhere and piece-wise, which is the case when $f$ is continuous and $u$ is piece-wise continuous) we have :
	\begin{equation}
		V^\pi(\zeta_t, t) = d(\delta_t) + \ln(\gamma) J^\pi(\zeta_t, t) \label{eq:second_prob}
	\end{equation}
	with $ln(\gamma) < 0$. We can then conclude the proof.
\end{proof}

We have reformulated the problem in an equivalent way using the classical variables of reinforcement learning, which are the reward and the value function. This property has already been proved in the discrete case \citep{Keramati}. It establishes a link between the maximization of the integral of the discounted rewards and the minimization of the integral of the discounted drive.

\subsubsection{Properties of the reward and the drive function}

In this section, we take the derivative of the reward function with respect to several quantities (realized in discrete time in Keramati), and study the sign to show the underlying properties of this function, reflecting behaviors in the agent.

We define the drive as
\begin{equation}
	d(\delta)= \sqrt{\delta^T \delta}
	\label{equation:drive}
\end{equation}
(in practice, $\sqrt{\epsilon + \delta^T \delta}$ to take the derivative in $0$). The reward at time $t$ is therefore
\begin{equation}
	r(t) = -(d(\delta))'(t) = - \delta_{t}^{T} \dot{\delta_{t}} / \sqrt{\delta_{t}^{T} \delta_{t}}
\end{equation}

We consider a situation in which an agent starts at time $t_0 = 0$ with a state $\delta_0 = [\delta_{0,1}, \delta_{0,2}, \dots]^T$, where $\delta_{0,1}$ and $\delta_{0,2}$ represent the levels of the agent's first two needs. From $t_0$ onwards, the agent continuously consumes the same resource which gives it a control $u = [m, 0, \dots, 0]$ constant in time, with $m$ the quantity of resource consumed per unit of time. Let us consider a time $t$ sufficiently close to $t_0$ so that the regulating effect of the body is negligible compared to the quantity of resource ingested. We have $\delta_t = \delta_0 + tu$ and the drive and the reward at time $t$ are

\begin{align}
	& d(t) = \sqrt{t^2m^2 + 2tm\delta_{0,1} + \delta_{0}^{T}\delta_{0}} \text{,} \\
	& r(t) = - \frac{(\delta_{0,1} + tm)m}{\sqrt{t^2m^2 + 2tm\delta_{0,1} + \delta_{0}^{T}\delta_{0}}}.
	\label{eq:drive_reward_derivative}
\end{align}

\textbf{Effects of deviation from the homeostatic set point for the feature receives an outcome:} Taking the derivative of the reward with respect to$|\delta_{0,1}|$, we find that

\begin{equation}
	\frac{\partial r(t)}{\partial |\delta_{0,1}|} \left\{
	\begin{aligned}
		\leq 0 & \text{ if } \delta_{0,1} \geq 0  \\ 
		\geq 0 & \text{ if } \delta_{0,1} \leq 0  \\ 
	\end{aligned}
	\right.
	.
	\label{derivee_1}
\end{equation}

The first case means that if an agent has exceeded its homeostatic set point for a need, and it continues to consume a resource affecting this need, it will receive a punishment (negative reward) that is proportional to the homeostatic setpoint deviation. The second case means that for an agent deprived of a resource, a fixed amount consumed of that resource will have a greater motivational outcome if the agent's initial need for the resource was high rather than low, as observed in \citep{Hodos}.

\textbf{Cross need interactions, effects of deviation from the homeostatic set point for a feature that does not receive an outcome:} Taking the derivative of the reward with respect to $|\delta_{0,2}|$, we find that

\begin{equation}
	\frac{\partial r(t)}{\partial |\delta_{0,2}|} \left\{
	\begin{aligned}
		\leq 0 & \text{ if } \delta_{0,1} + tm \leq 0  \\ 
		\geq 0 & \text{ if } \delta_{0,1} + tm \geq 0   \\ 
	\end{aligned}
	\right.
	.
	\label{derivee_2}
\end{equation}

In the first situation, $\delta_{0,1} + tm \leq 0$, so $x_{t,1} = x_{0,1} + tm \leq x^{*}_{1}$ and the agent is still below its homeostatic set point for the first need at time $t$. The agent will gain a positive reward by consuming the resource affecting its first need, but the negative derivative means that this reward will be reduced if $|\delta_{0,2}|$ increases. The interpretation of the second situation is similar, but the reward is now negative, since the agent has exceeded its homeostatic set point for the first need. Such inhibitory effects occur in nature, as shown experimentally by \citep{Dickinson}. For example, food deprivation tending to suppress water-related responses.

\textbf{Effects of resource dose:} Taking the derivative of drive with respect to $tm$, which is the amount of resource consumed at time $t$, we find that: 

\begin{equation}
	\frac{\partial d(t)}{\partial tm} \left\{
	\begin{aligned}
		\leq 0 & \text{ if } \delta_{0,1} + tm \leq 0  \\ 
		\geq 0 & \text{ if } \delta_{0,1} + tm \geq 0   \\ 
	\end{aligned}
	\right.
	.
	\label{derivee_3}
\end{equation}

This means that if an agent has not reached its homeostatic setpoint for a need at time $t$, i.e. $\delta_{0,1} + tm \leq 0$, then its training would have been smaller. On the other hand, if the agent has reached closer to its homeostatic setpoint $\delta_{0,1} + tm \geq 0 $, then its amount consumed is closer, as shown for rats in \citep{Skjoldager}.  \textbf{<Need clarity>} Next we present the results of the experiment and discuss them. 


\section{Results and Discussion}
\begin{figure}[H]
	\centering
	\subfigure[]{%
		\includegraphics[width=3.5cm, height=3.5cm]{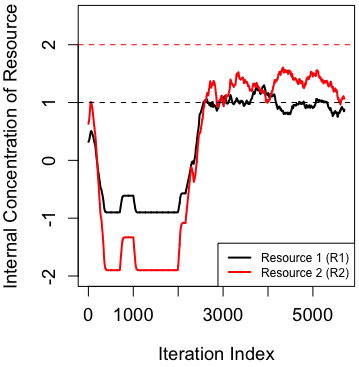}
		\label{fig:sim_1}}
	\subfigure[]{
		\hspace{0.225cm}
		\includegraphics[width=3.5cm, height=3.5cm]{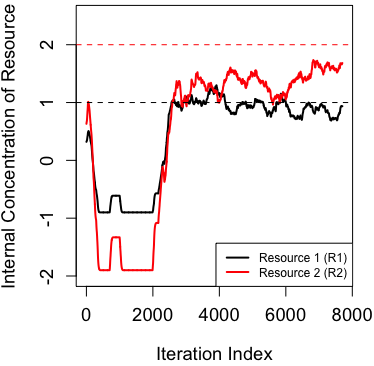}
		\label{fig:sim_2}}
	\subfigure[]{%
		\includegraphics[width=3.5cm, height=3.5cm]{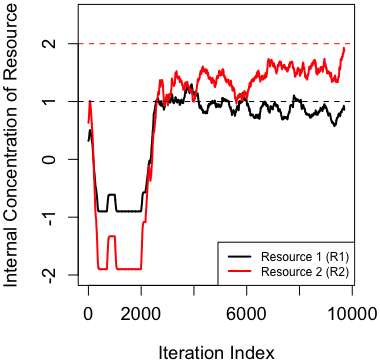}
		\label{fig:sim_3}}
	\subfigure[]{%
		\hspace{-0.225cm}
		\includegraphics[width=3.5cm, height=3.5cm]{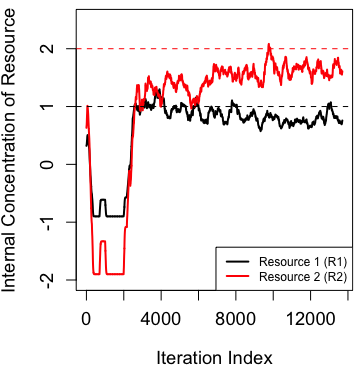}
		\label{fig:sim_4}}
	\caption{Resource consumption for the two resources in the square environment. The homeostatic set point for Resource 1 is 1 and for Resource 2 is 2, as indicated by dashed black and red lines respectively. (a) : 6000 iterations (b): 8000 iterations. (c): 10000 iterations (d):14000 iterations. }
	\label{fig:resource_consumption2}
\end{figure}

\begin{figure}[]
	\centering
	\includegraphics[width=5cm]{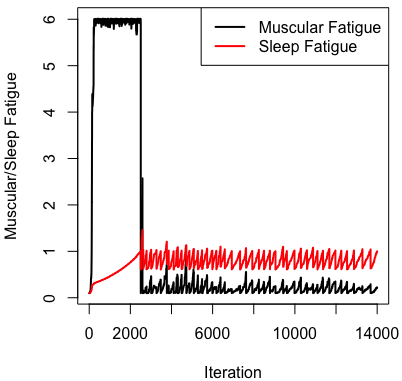}
	\caption{Plot showing the variation in the muscular and sleep fatigue.}
	\label{fig:muscle_sleep_fatigue}
\end{figure}

Figure ~\ref{fig:resource_consumption2} shows the results of resource concentration for the agent w.r.t time (i.e. iterations). In the initial stages, the agent begins with very limited amount of both $R_1$ and $R_2$, and as it explores the environment, its energy decreases as is indicated by the zero and later negative concentration for the resources (shock state). Corresponding to this, the muscular fatigue and sleep fatigue also increases for the agent (Figure ~\ref{fig:muscle_sleep_fatigue}).

 Gradually, with environmental exploration the agent is reaches the resource reservoirs and registers the changes in its internal state due to this exploratory action. As a result of this, the agent slowly learns the actions that lead to the resources given its internal state and position. In the Figure ~\ref{fig:sim_1}, which graphs the change in the internal concentration of the resources for 6000 iterations, it is observed that the agent begins to marginally achieve the homeostasis for $R_1$. 
 By 8000 iterations (Figure ~\ref{fig:sim_2}), the agent has also started to approach its homeostatic set point for Resource $R_2$ (red line). This behaviour reinforces the agent's intelligence related to homeostatic deviation from $R_2$. The agent strives for $R_2$ but in the process, the concentration of $R_1$ is also maintained closer to its homeostatic set point. Thus, it appears that the agent intends to learn a policy that leads to global homeostasis. 
  
 To verify whether the observed graphs in Figures ~\ref{fig:sim_1} and ~\ref{fig:sim_2} are results or purely exploratory or exploitative actions, we tested the algorithm for 10000 (Figure ~\ref{fig:sim_3})  and 14000 (Figure ~\ref{fig:sim_4}) iterations. These graphs confirm the learning process for the agent, and we observe that no matter how long the iterations last, the resource concentrations lie close to their respective homeostatic set points.  By, 14000 iteration the agent begins to achieve a plateau in its behaviour, which reflects that the agent has learned to directly leap (walk in a directed manner) to the resource reservoirs in times of homeostatic deviation. Note that the final concentrations for Resources $R_1$ and $R_2$ are much higher than the initial starting point, confirming that the agent has learnt to take actions that lead to homeostasis. 

Overall, it is observed that 70\% actions taken are exploitative and 30\% exploratory over the life-course of the agent for each iteration case. The change in muscular and sleep fatigue for the agent as its learns to achieve homeostasis in an unknown environment is shown in Figure ~\ref{fig:muscle_sleep_fatigue}. As mentioned before, when the agent starts its exploration, the muscular fatigue rises very fast. Gradually, as the agent learns to identify resource position and consumes the resource, both muscle and sleep fatigue reduce to minimum values. In fact, both kinds of fatigue achieve relative stability after 2000 iteration.

\begin{figure}[h]
	\centering
	\subfigure[]{%
		\includegraphics[width=3.5cm, height=3.5cm]{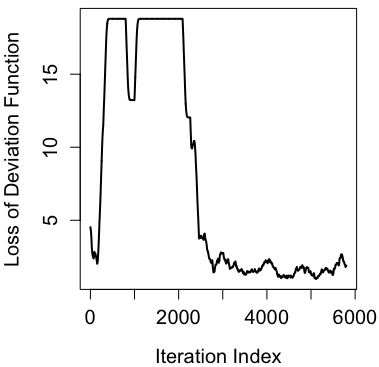}
		\label{fig:loss_1}}
	\subfigure[]{
		\includegraphics[width=3.5cm, height=3.5cm]{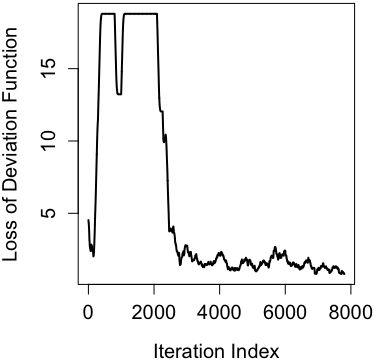}
		\label{fig:loss_2}}
	\subfigure[]{%
		\includegraphics[width=3.5cm, height=3.5cm]{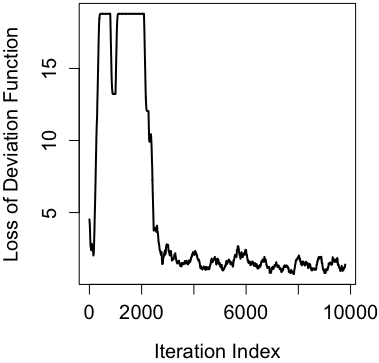}
		\label{fig:loss_3}}
	\subfigure[]{%
		\includegraphics[width=3.5cm, height=3.5cm]{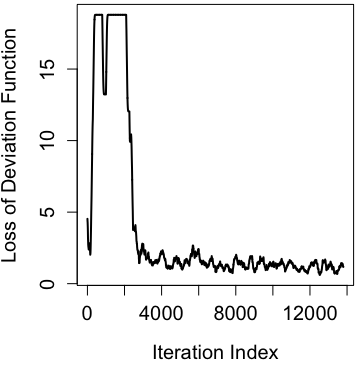}
		\label{fig:loss_4}}
	\caption{Plots showing the variation in the Loss of Deviation Function ($J$). (a) : 6000 iterations (b): 8000 iterations. (c): 10000 iterations (d):14000 iterations.}
	\label{fig:resource_consumption}
\end{figure}

We also show the graph of Deviation Function loss w.r.t iterations (Figure ~\ref{fig:resource_consumption}). These plots show that in the beginning of the task, the agent's homeostatic deviation increases because it is exploring an environment without too much resources in its system. This is also seen in Figure ~\ref{fig:resource_consumption2}.  After 3000$^{th}$ iteration, the resource concentration does not fall below 1 for each of the resources, indicating that learning has started to become concrete in the agent. As the iterations increase, agent learning gets solidified and the agent can immediately go to the resource points to satiate itself, without having to explore the environment too much. After 2000 iterations, the Loss of Deviation function begins to decline, gradually approaching zero as iterations increase. 

\begin{figure*}[h]
	\centering
	\subfigure[]{
		\includegraphics[width=3.5cm, height=3.5cm]{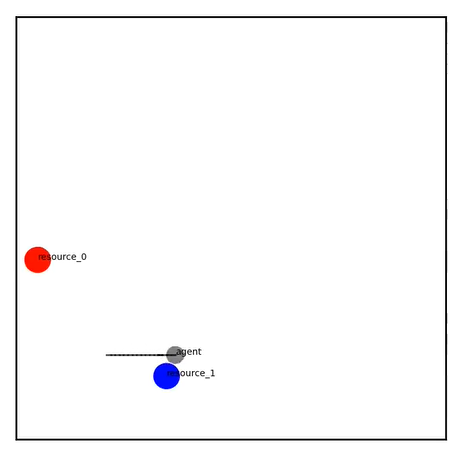}
		\label{fig:1}}
	\subfigure[]{%
		\includegraphics[width=3.5cm, height=3.5cm]{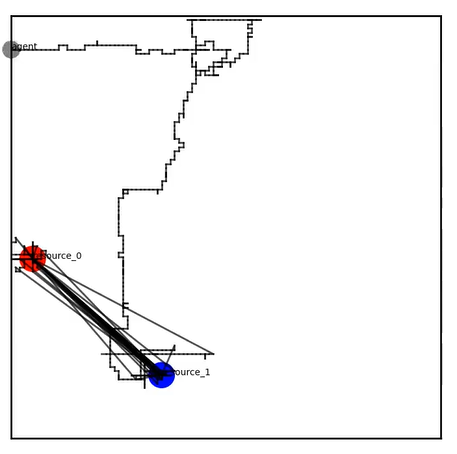}
		\label{fig:2}}
	\subfigure[]{
		\includegraphics[width=3.5cm, height=3.5cm]{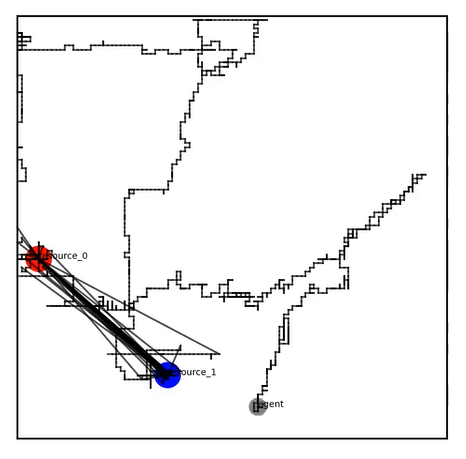}
		\label{fig:3}}
	\subfigure[]{
		\includegraphics[width=3.5cm, height=3.5cm]{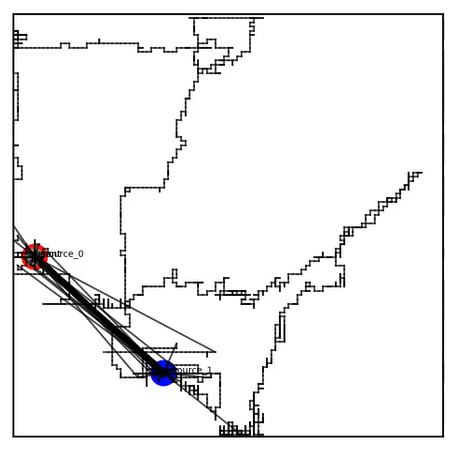}
		\label{fig:4}}
		\subfigure[]{
		\includegraphics[width=3.5cm, height=3.5cm]{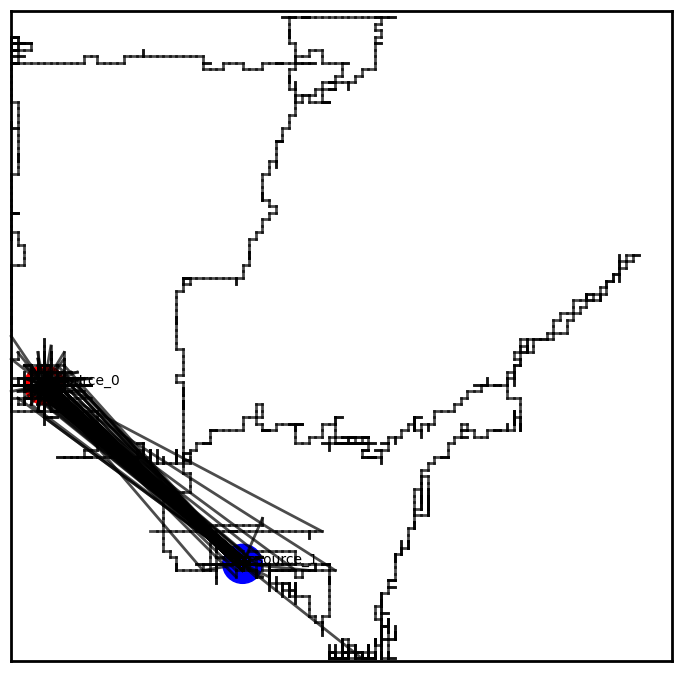}
		\label{fig:5}}
	\subfigure[]{
		\includegraphics[width=3.5cm, height=3.5cm]{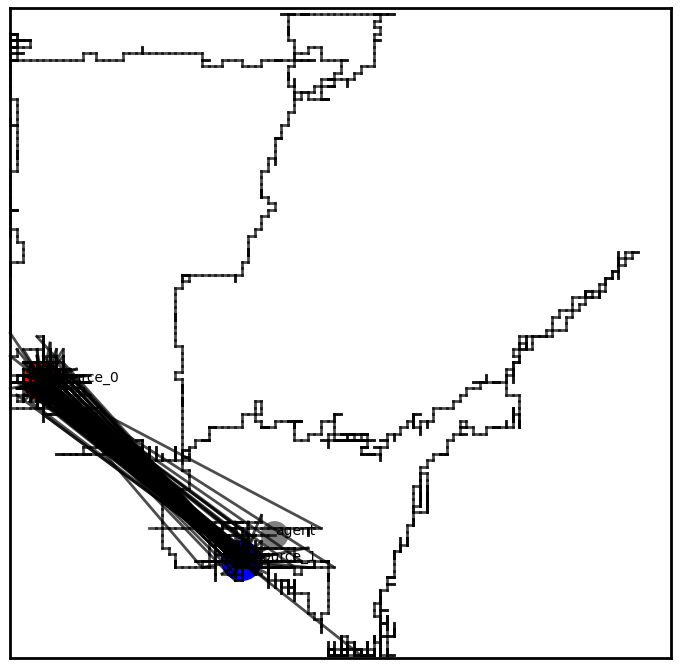}
		\label{fig:6}}
	\subfigure[]{
		\includegraphics[width=3.5cm, height=3.5cm]{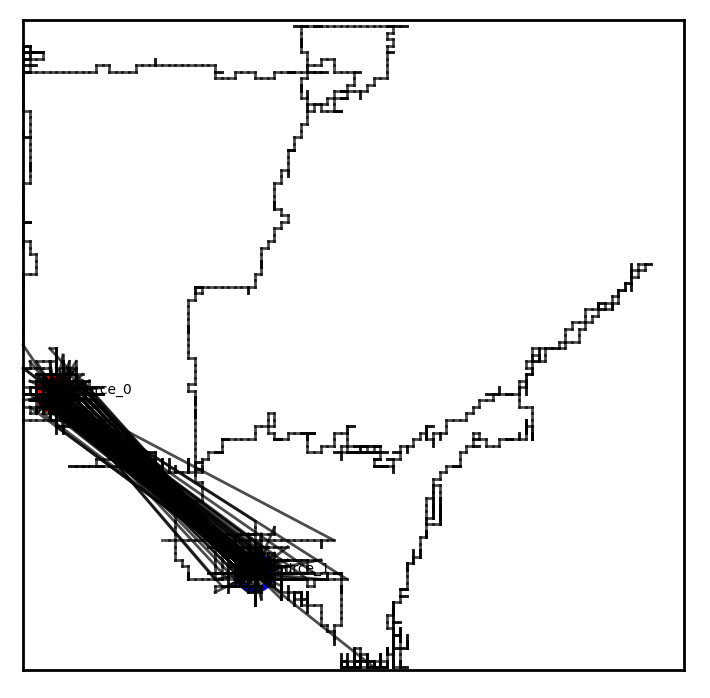}
		\label{fig:7}}
	\subfigure[]{
		\includegraphics[width=3.5cm, height=3.5cm]{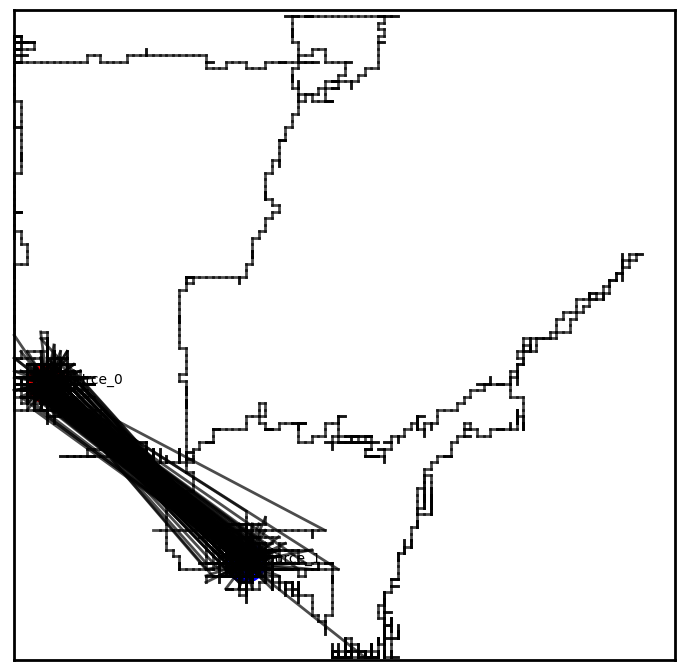}
		\label{fig:8}}
	\label{fig:plots}
	\caption{Agent Learning and Exploration in an unknown environment. The figure shows agent track for the duration of iteration. (a): Beginning point (b) : Exploring the environment (c): Further exploration. (d): Learning Resource positions. (e) 6000 iteration. (f) 8000 iterations. (g) 10000 iterations. (h) 14000 iterations}\label{fig:memory}
\end{figure*}

Figure ~\ref{fig:plots} shows the trajectory of the agent movement in the environment as determined by the actions taken in each iteration. Exploration is seen as the trajectory away from the resource points, until by Figure ~\ref{fig:5}, the agent has learnt the resource positions and is able to consume them by directly walking to the resources. The line joining the two resources gradually becomes darker (Exploitation) as opposed to trajectories beyond these positions (Exploration). The dark pathways in the graphs testify the policy learnt by the agent. 

 Finally, as per the neural network based algorithm designed by us, the agent reaches an optimal point and its learning is solidified in this environment such that when the agent senses resource depletion, due to normal bodily processes, it is able to replenish itself quickly having learned the resource positions. Thus, the agent can dynamically maintain homeostasis in real world in a  continuous-time manner. 

 It is vital to note that as the biological process of homeostasis never ends, in the simulations also, our program is never ending and continues indefinitely as seen in Figure ~\ref{fig:6}. But, we closely observe the agent's behaviour for multiple iterations far apart : 6000, 8000, 10000 and 14000. This allows us to observe and study the long-term behaviour of the agent and discover any anomalies in the agent's behaviour pertaining to homeostatic regulation. The end result of this experiment is, a solidified neural trace/path (Figure ~\ref{fig:6}) that is reflective of an established learning (policy) in the given environment. The complete simulations are shared on Github \citep{hrrl_code}. 

It is vital to discuss our results in the context of other studies. \citeauthor{yoshida2021embodiment} proposed a neural homeostat in which the agent stabilises its internal physiology through interaction with the environment \citep{yoshida2021embodiment}. Authors considered two kinds of homeostasis: primary \textit{homeostasis} relating to direct internal body control and \textit{behavioural homeostasis} which entails change in agent behaviour (drinking water, eating food) depending on its interaction with the environment. In our work we do not distinguish homeostasis as internal or external because the external behaviour and internal deviation are closely linked to each other. The internal deviations lead to external actions, which in turn affect the internal state or physiology of the agent. We consider agent-environment interaction as a separate phenomenon through which the agent learns about the environment boundaries and gradually about the resource points. 

\citeauthor{yoshida2021embodiment} also consider three types of information received by the agent : exteroception, proprioception, and interoception and compute total reward as the sum of homeostatic reward and proprioceptive cost. Essentially, the reward is defined by coupling the agent's internal state dynamics and the environment. The authors find that simple food-capturing reward does not result in homeostatic behaviour. This is in contrast to our work wherein reward is modeled as sum of both immediate and long-term reward, and the agent is able to achieve homeostatic behavior and learning. Furthermore, the agent we considered was guided only be interoception and through the exploration of the environment could learn about information from the environment and resource points. Thus, no exteroception was used in our work. In a different work, \citeauthor{Yoshida2016OnRF} formulates homeostasis achievement as a survival problem, and presents maximization of the multi-step survival probability as the solution. In the most recent work, \citeauthor{Yoshida2023HomeostaticRL} attempted to increase learning speed of the homeostatic RL agent, by introducing interoceptive soft-behaviour switching in the algorithm.  

Another difference in our work from contemporary research is the focus on monolithic agent for a classical \textit{two-resource problem}. The concept of modular agent is proposed by \citeauthor{Dulberg2023HavingS} according to which each sub-agent achieves the homeostasis through divide and conquer approach. Unlike the global optimization problem in the monolithic agent, this work emphasizes the use of decentralized control which is rooted in adaptation rather than centralized control. Authors demonstrate that such a modular approach is able to solve the challenge of multi-dimensional homeostatic regulation and each sub-network corresponding to the sub-agent learn distinct policies based on separate reward components. \citeauthor{ouka} confirm the applicability of HRRL for foraging strategies. Authors considered three foraging rules that the agent can use depending upon their environment : Closest Distance (CD), No  Interaction (NI) and Equal Distance (ED). For each of these they suggested different rules for internal state update. However, it is unclear how the agent will decide which foraging strategy to use. Moreover, in their experiment, the episode terminated if the internal states deviated from the homeostatic set point. In contrast, in our experiment the agent continued to explore the environment and forage to find and consume resources required to maintain homeostasis. 

Our work is partially similar to that of Walter \citep{Lettvin1954TheLB} in which robotic agents have to recharge themselves by searching for the batteries at the recharge stations scattered in an environment. However, in our case, we use only one agent and the aim is to mimic an autonomous biological agent. In our work we demonstrate resource foraging behaviour while also accounting for muscular and sleep fatigue. Explaining embodied behaviours aligned with the physiology of the biological agent may be more complex and complicated, as these may not be explicitly associated with organism survival. For e.g. pro-social behaviours, desire for recognition, gambling etc.  Nevertheless, it is possible that these behaviors are somehow translated, on a small scale, into a set of characteristics that could be represented in terms of $x$ motor function variables. Indeed, \citeauthor{Juechems} argue that even non-physiological motivations can be modeled using the HRRL framework. Seeking long-term goal is an example of this, wherein intermediate goals lead to the final goal. Thus, there appears to be a structural and conceptual similarity between the learning mechanism for complex goals and primitive goals for homeostasis. 

Finally, our model is essentially limited by knowledge of the human body and the structure of more abstract needs, which means that defining specific training functions for complex goals is a challenge. HRRL is also challenged by the number of internal variables that can become very large, making it difficult for the algorithm to converge and thus for the agent to learn. A research goal could be to model robots possessing automatically learned human characteristics, which can evolve and interact together in an environment. A guiding example is discussed by \citeauthor{Dulberg2023HavingS}. Despite its limitations, the simplicity of our model and its ability to have an arbitrary choice of scale may have an impact on this goal. We have made simplifications on the generality of the simulation, notably on the lack of stochasticity by not putting other agents in the environment, but we could in the future make simulations with several agents including prey and predators. In a multi-agent simulation, we could also create a colony of agents in which each member implements de facto empathy for its cohorts, as suggested by \citep{Man}.



\section{Conclusion}
We extended the HRRL framework in continuous time space by developing an agent capable of dynamic and long-term homeostatic regulation. Basic properties of food foraging in an unknown environment, and physiological attributes like muscle and sleep fatigue were embodied in the agent. Through computer simulations we demonstrated that the agent was able learn to select action that lead to homeostasis in an unknown environment which contained the resources necessary for its survival. The characteristics of muscle and sleep fatigue were also discussed as the agent learnt to achieve homeostasis. Aligned with the unending biological process of homeostasis our simulation results showed that the agent continued its learnt policy to maintain homeostasis for the two resources. Finally, we call this framework in continuous-time and space, as the Continuous-Time Continuous-Space HRRL : CTCS-HRRL.

\section*{Acknowledgement}

This work has been supported by ANR-17-EURE-1553-0017, and ANR-10-IDEX-0001-02. BSG acknowledges funding from the Basic Research Program at the National Research University Higher School of Economics (HSE University). The funders had no role in study design, data collection and analysis, decision to publish, or preparation of the manuscript.


\vskip 0.2in
\bibliography{biblio}

\begin{thebibliography}{40}
\providecommand{\natexlab}[1]{#1}
\providecommand{\url}[1]{\texttt{#1}}
\expandafter\ifx\csname urlstyle\endcsname\relax
  \providecommand{\doi}[1]{doi: #1}\else
  \providecommand{\doi}{doi: \begingroup \urlstyle{rm}\Url}\fi

\bibitem[Arikawa et~al.(2023)Arikawa, Yoshida, and Kanazawa]{ouka}
Etsushi Arikawa, Naoto Yoshida, and Hoshinori Kanazawa.
\newblock Homeostatic reinforcement learning explains foraging strategies.
\newblock \emph{11th International Symposium on Adaptive Motion of Animals and
  Machines (AMAM2023)}, pages 160--161, 2023.

\bibitem[Bhargava(2023)]{hrrl_code}
Yesoda Bhargava.
\newblock Hrrl simulations {GitHub}, 2023.
\newblock URL \url{https://github.com/vagansh/HRRL/tree/main/Simulations}.

\bibitem[Dickinson and Balleine(2002)]{Dickinson}
A.~Dickinson and B.~W. Balleine.
\newblock The role of learning in motivation.
\newblock \emph{Volume 3 of Steven’s Handbook of Experimental Psychology:
  Learning, Motivation, and Emotion}, 3:\penalty0 497--533, 2002.

\bibitem[Doya(2000)]{Doya}
K.~Doya.
\newblock Reinforcement learning in continuous time and space.
\newblock \emph{Neural computation}, 12(1):\penalty0 219--245, 2000.
\newblock \doi{10.1162/089976600300015961}.

\bibitem[Dulberg et~al.(2023)Dulberg, Dubey, Berwian, and
  Cohen]{Dulberg2023HavingS}
Zachary Dulberg, Rachit Dubey, Isabel~M. Berwian, and Jonathan~D. Cohen.
\newblock Having “multiple selves” helps learning agents explore and adapt
  in complex changing worlds.
\newblock \emph{bioRxiv}, 2023.
\newblock URL \url{https://api.semanticscholar.org/CorpusID:254879274}.

\bibitem[Hodos(1961)]{Hodos}
W.~Hodos.
\newblock Progressive ratio as a measure of reward strength.
\newblock \emph{Science}, 134:\penalty0 943--944, 1961.
\newblock \doi{10.1126/science.134.3483.943}.

\bibitem[Hull(1943)]{Hull}
Clark Hull.
\newblock Principles of behavior.
\newblock \emph{Appleton Century}, 1943.

\bibitem[Juechems and Summerfield(2019)]{Juechems}
Keno Juechems and Christopher Summerfield.
\newblock Where does value come from?
\newblock \emph{Preprint}, 2019.
\newblock \doi{10.31234/osf.io/rxf7e}.

\bibitem[Keramati(2013)]{KeramatiThesis}
Mehdi Keramati.
\newblock A homeostatic reinforcement learning theory and its implications in
  cocaine addiction.
\newblock \emph{PhD thesis}, 2013.

\bibitem[Keramati and Gutkin(2014)]{Keramati2014}
Mehdi Keramati and Boris Gutkin.
\newblock Homeostatic reinforcement learning for integrating reward collection
  and physiological stability.
\newblock \emph{Elife}, 2014.
\newblock \doi{10.7554/eLife.04811}.

\bibitem[Keramati and Gutkin(2011)]{Keramati}
Mehdi Keramati and Boris~S. Gutkin.
\newblock A reinforcement learning theory for homeostatic regulation.
\newblock \emph{Advances in Neural Information Processing Systems},
  24:\penalty0 82--90, 2011.
\newblock \doi{10.5555/2986459.2986469}.

\bibitem[Konidaris and Barto(2006)]{Konidaris}
George Konidaris and Andrew~G. Barto.
\newblock An adaptive robot motivational system.
\newblock \emph{From Animals to Animats 9, 9th International Conference on
  Simulation of Adaptive Behavior}, pages 346--356, 2006.
\newblock \doi{10.1007/11840541_29}.

\bibitem[Kretchmar(2000)]{Kretchmar}
R.~Matthew Kretchmar.
\newblock A synthesis of reinforcement learning and robust control theory.
\newblock \emph{PhD thesis}, 2000.

\bibitem[Kriegeskorte and Douglas(2018)]{Kriegeskorte}
Nikolaus Kriegeskorte and Pamela~K. Douglas.
\newblock Cognitive computational neuroscience.
\newblock \emph{Nature Neuroscience}, 21:\penalty0 1148--1160, 2018.
\newblock \doi{10.1038/s41593-018-0210-5}.

\bibitem[Krizhevsky et~al.(2012)]{Krizhevsky}
Alex Krizhevsky et~al.
\newblock Imagenet classification with deep convolutional neural networks.
\newblock \emph{Advances in Neural Information Processing Systems},
  25:\penalty0 1097--1105, 2012.
\newblock \doi{10.1145/3065386}.

\bibitem[Lettvin(1954)]{Lettvin1954TheLB}
Jerome~Y. Lettvin.
\newblock The living brain: W. grey walter. w.w. norton \& co., inc., new york,
  1953, 311 pp., \$3.95.
\newblock \emph{Electroencephalography and Clinical Neurophysiology},
  6:\penalty0 353--354, 1954.
\newblock URL \url{https://api.semanticscholar.org/CorpusID:141480422}.

\bibitem[Lussange et~al.(2020)]{Lussange}
Johann Lussange et~al.
\newblock Modelling stock markets by multi-agent reinforcement learning.
\newblock \emph{Comput Econ}, 2020.
\newblock \doi{10.1007/s10614-020-10038-w}.

\bibitem[Man and Damasio(2019)]{Man}
K.~Man and A.~Damasio.
\newblock Homeostasis and soft robotics in the design of feeling machines.
\newblock \emph{Nat Mach Intell}, 1:\penalty0 446--452, 2019.
\newblock \doi{10.1038/s42256-019-0103-7}.

\bibitem[Matignon(2006)]{Matignon}
Laëticia Matignon.
\newblock Reward function and initial values : Better choices for accelerated
  goal-directed reinforcement learning.
\newblock \emph{Lecture Notes in Computer Science, Springer}, 1(4131):\penalty0
  840--849, 2006.
\newblock \doi{doi.org/10.1644/BHE-004.1}.

\bibitem[Mnih et~al.(2013)]{Mnih}
Volodymyr Mnih et~al.
\newblock Playing atari with deep reinforcement learning.
\newblock \emph{Preprint 1312.5602}, 2013.

\bibitem[Niv(2009)]{Niv}
Yael Niv.
\newblock Reinforcement learning in the brain.
\newblock \emph{Journal of Mathematical Psychology}, 53:\penalty0 139--154,
  2009.
\newblock \doi{10.1016/j.jmp.2008.12.005}.

\bibitem[Panerai et~al.(1999)]{Panerai}
R.~B. Panerai et~al.
\newblock Effect of carbon dioxide on dynamic cerebral autoregulation
  measurement.
\newblock \emph{Physiological Measurement}, 20:\penalty0 265--275, 1999.
\newblock \doi{10.1088/0967-3334/20/3/304}.

\bibitem[Pattaranit and van~den Berg(2008)]{Pattaranit}
Ratchada Pattaranit and Hugo~Antonius van~den Berg.
\newblock Mathematical models of energy homeostasis.
\newblock \emph{J. R. Soc. Interface}, 5:\penalty0 1119--1135, 2008.
\newblock \doi{10.1098/rsif.2008.0216}.

\bibitem[Polynikis et~al.(2009)Polynikis, Hogan, and di~Bernardo]{Polynikis}
A~Polynikis, S~J Hogan, and M~di~Bernardo.
\newblock Comparing different ode modelling approaches for gene regulatory
  networks.
\newblock \emph{J Theor Biol}, 261(4):\penalty0 511--530, 2009.
\newblock \doi{10.1016/j.jtbi.2009.07.040}.

\bibitem[Ramsay and Woods(2014)]{Ramsay}
Douglas Ramsay and Stephen Woods.
\newblock Clarifying the roles of homeostasis and allostasis in physiological
  regulation.
\newblock \emph{Psychol Rev}, 121(2):\penalty0 225--247, 2014.
\newblock \doi{10.1037/a0035942}.

\bibitem[Richards(2019)]{Richards}
Blake~A. Richards.
\newblock A deep learning framework for neuroscience.
\newblock \emph{Nature Neuroscience}, 22:\penalty0 1761--1770, 2019.
\newblock \doi{10.1038/s41593-019-0520-2}.

\bibitem[Shteingart and Loewenstein(2014)]{Shteingart}
Hanan Shteingart and Yonatan Loewenstein.
\newblock Reinforcement learning and human behavior.
\newblock \emph{Current Opinion in Neurobiology}, 25:\penalty0 93--98, 2014.
\newblock \doi{10.1016/j.conb.2013.12.004}.

\bibitem[Silver et~al.(2016)]{Silver}
David Silver et~al.
\newblock Mastering the game of go with deep neural networks and tree search.
\newblock \emph{Nature}, 529:\penalty0 484--489, 2016.
\newblock \doi{10.1038/nature16961}.

\bibitem[Silver et~al.(2018)]{Silver2018}
David Silver et~al.
\newblock A general reinforcement learning algorithm that masters chess, shogi,
  and go through self-play.
\newblock \emph{Science}, 362(6419):\penalty0 1140--1144, 2018.
\newblock \doi{10.1126/science.aar6404}.

\bibitem[Skjoldager et~al.(1993)Skjoldager, Pierre, and Mittleman]{Skjoldager}
P.~Skjoldager, P.~J. Pierre, and G.~Mittleman.
\newblock Reinforcer magnitude and progressive ratio responding in the rat:
  Effects of increased effort, prefeeding, and extinction.
\newblock \emph{Learn motiv}, 24(3):\penalty0 303--343, 1993.
\newblock \doi{10.1006/lmot.1993.1019}.

\bibitem[Staddon(1983)]{Staddon}
J.~E.~R. Staddon.
\newblock Adaptive behavior and learning.
\newblock \emph{Cambridge University Press}, 1983.

\bibitem[Sutton and Barto(2018)]{Sutton}
Richard~S. Sutton and Andrew~G. Barto.
\newblock Reinforcement learning: An introduction.
\newblock \emph{http://incompleteideas.net/book/bookdraft2018jan1.pdf}, 2018.

\bibitem[Tallec et~al.(2019)Tallec, Blier, and Ollivier]{Tallec}
Corentin Tallec, Léonard Blier, and Yann Ollivier.
\newblock Making deep q-learning methods robust to time discretization.
\newblock \emph{Preprint 1901.09732}, 2019.

\bibitem[Thomson and Blantz(2008)]{Thomson}
Scott~C. Thomson and Roland~C. Blantz.
\newblock Glomerulotubular balance, tubuloglomerular feedback, and salt
  homeostasis.
\newblock \emph{JASN}, 19:\penalty0 2272--2275, 2008.
\newblock \doi{10.1681/ASN.2007121326}.

\bibitem[Toates(1986)]{Toates}
F.~M. Toates.
\newblock Motivational systems (problems in the behavioral sciences).
\newblock \emph{Cambridge University Press}, 1986.

\bibitem[Versypt et~al.(2015)Versypt, Makrides, et~al.]{Versypt}
Ashlee N.~Ford Versypt, Elizabeth Makrides, et~al.
\newblock Bifurcation study of blood flow control in the kidney.
\newblock \emph{Mathematical biosciences}, 263:\penalty0 169--179, 2015.
\newblock \doi{10.1016/j.mbs.2015.02.015}.

\bibitem[Wingfield(2005)]{Wingfield}
John Wingfield.
\newblock The concept of allostasis: Coping with a capricious environment.
\newblock \emph{Journal of Mammalogy}, 86(2):\penalty0 248--254, 2005.
\newblock \doi{doi.org/10.1644/BHE-004.1}.

\bibitem[Yoshida(2016)]{Yoshida2016OnRF}
Naoto Yoshida.
\newblock On reward function for survival.
\newblock \emph{ArXiv}, abs/1606.05767, 2016.
\newblock URL \url{https://api.semanticscholar.org/CorpusID:9947618}.

\bibitem[Yoshida et~al.(2021)Yoshida, Daikoku, Nagai, and
  Kuniyoshi]{yoshida2021embodiment}
Naoto Yoshida, Tatsuya Daikoku, Yukie Nagai, and Yasuo Kuniyoshi.
\newblock Embodiment perspective of reward definition for behavioural
  homeostasis.
\newblock In \emph{Deep RL Workshop NeurIPS 2021}, 2021.
\newblock URL \url{https://openreview.net/forum?id=kG_4YfvbCJo}.

\bibitem[Yoshida et~al.(2023)Yoshida, Kanazawa, and
  Kuniyoshi]{Yoshida2023HomeostaticRL}
Naoto Yoshida, Hoshinori Kanazawa, and Yasuo Kuniyoshi.
\newblock Homeostatic reinforcement learning through soft behavior switching
  with internal body state.
\newblock \emph{2023 International Joint Conference on Neural Networks
  (IJCNN)}, pages 1--8, 2023.
\newblock URL \url{https://api.semanticscholar.org/CorpusID:260386175}.

\end{thebibliography}

\section*{Appendix}

\section{Details of the realization of the experiment}

\subsection{Parameters of the environment}

Minimum length between two "corners" of the environment: $1$ unit. 
\newline
Radius of the circles within which the agent can consume a resource: $0.3$ unit. 
\newline
Definition of the features of the internal state of the agent at time $t$: $x_t = [r_{1,t}, r_{2,t}, f_{m,t}, f_{s,t}]$ where $r_{i,t}$ is the $i$-th feature associated with the $i$-th resource necessary for the survival of the agent, $f_{m,t}$ is the muscular fatigue and $f_{s,t}$ is the sleep-related fatigue.
\newline
Homeostatic set point for the features of the internal state of the agent: $x^{*} = [1,2,0,0]$.
\newline
Definition of the features of the environment of the agent at time $t$: $e_t = [x_{t}, y_{t}]$ where $x_{t}$ is the x-axis coordinate for the agent and $y_{t}$ is the y-axis coordinate. 
\newline
Definition of the features of the state of the agent at time $t$: $\zeta_t = [r_{1,t}, r_{2,t}, f_{m,t}, f_{s,t}, x_{t}, y_{t}]$.
\newline
Control associated with the action of walking left: $u = [0,0,0.01,0,-0.1,0]$.
\newline
Control associated with the action of walking right: $u = [0,0,0.01,0,0.1,0]$.
\newline
Control associated with the action of walking down: $u = [0,0,0.01,0,0,-0.1]$.
\newline
Control associated with the action of walking up: $u = [0,0,0.01,0,0, 0.1]$.
\newline
Constraint associated with the action of walking: The agent can only take this action if $f_{m,t} \leq 6$ and if it does not leave its environment by performing it.
\newline
Control associated with the action of going to resource: $u = [0,0,0.01,0,0,0]$.
\newline
Control associated with the action of sleeping (constant in time for the entire duration of sleep): $u = [0,0,0,-0.001,0,0]$.
\newline
Constraint associated with the action of sleeping: The agent can only take the action of sleeping if $f_{s,t} \geq 1$. It must then sleep for a minimum time equivalent to 1000 times the elementary time, and cannot take any other action during this time. At the end, it can choose to resume the action of sleeping or perform another action. If $f_{s,t} \geq 10$, the only possible action for the agent is to sleep.
\newline
Control associated with the action of consuming the resource 1: $u = [0.1,0,0,0,0,0]$.
\newline
Control associated with the action of consuming the resource 2: $u = [0,0.1,0,0,0,0]$.
\newline
Constraint associated with the action of consuming a resource: The agent can only take this action if it is within the circle in which the resource is located and if $\zeta_{i,t} \leq 8$. When an agent sees a resource, which is in its visual field and at a distance of less than 4, it can choose the action equivalent to the succession of elementary actions leading it to this resource.
\newline
Control associated with the action of not doing anything: $u = [0,0,0,0,0,0]$.
\newline
The self-regulation function $f$ is defined as
\begin{equation}
	\dot \zeta = f(\zeta, u) =
	\begin{pmatrix}
		c_1 (\zeta_1 + x_{1}^{*}) + u_1 (\zeta_1 + x_{1}^{*}) \\
		c_2 (\zeta_2 + x_{2}^{*}) + u_2 (\zeta_2 + x_{2}^{*}) \\
		c_3 (\zeta_3 + x_{3}^{*}) + u_3 (\zeta_3 + x_{3}^{*}) \\
		c_4 (\zeta_4 + x_{4}^{*}) + u_4 (\zeta_4 + x_{4}^{*}) \\
		u_7 \\
		u_8 \\
	\end{pmatrix} 
\end{equation}
with $(c_1, c_2, c_3, c_4) = (-0.05, -0.05, -0.008, 0.0005)$.

\subsection{Parameters of the algorithm}

\textbf{General parameters}
\newline
The time step $\Delta_t$: 0.01.
\newline
The probability of selecting a random action $\epsilon$: 0.3.
\newline
The discount factor $\gamma$: 0.99.
\newline
The hyperparameter controlling the target function $\tau$: 0.001.
\newline
\textbf{Parameters of the neural networks}
\newline
Number of hidden layers of the neural networks: $2$. 
\newline
Number of neurons in the hidden layers: $128$. 
\newline
Dropout rate: $0.15$. 
\newline
Activation functions: Sigmoid, to make the network continuous, so it is easier to take the derivative of the deviation function $J$ with respect to the inputs.
\newline
Optimizer : Adam (with default parameters).
\newline
Learning rate: $0.001$. 

\end{document}